%% file: main_camera_ready.tex
\documentclass[runningheads]{llncs}

\usepackage{eccv}

\usepackage{eccvabbrv}

\usepackage{graphicx}
\usepackage{booktabs}

\usepackage[accsupp]{axessibility}  

\usepackage{floatrow}
\usepackage{comment}
\usepackage{bbding}
\usepackage{tabularx}
\usepackage{booktabs}
\usepackage{colortbl}
\usepackage{multirow}
\usepackage{xspace}
\usepackage[acronym]{glossaries}
\usepackage{float}
\usepackage{wrapfig}
\usepackage{algorithm}
\usepackage{algpseudocode}
\algtext*{EndWhile}
\algtext*{EndIf}
\algtext*{EndFor}

\usepackage{bbding}
\usepackage{pifont}

\usepackage{hyperref}

\usepackage{orcidlink}

\begin{document}

\title{TC4D: Trajectory-Conditioned\\Text-to-4D Generation}


\author{Sherwin Bahmani\inst{*1,2,3} \and
Xian Liu\inst{*4} \and
Wang Yifan\inst{*5} \and
Ivan Skorokhodov\inst{3} \and \\
Victor Rong\inst{1,2} \and
Ziwei Liu\inst{6} \and
Xihui Liu\inst{7} \and
Jeong Joon Park\inst{8} \and
Sergey Tulyakov\inst{3} \and \\
Gordon Wetzstein\inst{5} \and
Andrea Tagliasacchi\inst{1,9,10} \and
David B. Lindell\inst{1,2}
}

\authorrunning{Bahmani et al.}


\institute{$^{1}$University of Toronto\space$^{2}$Vector Institute\space$^{3}$Snap Inc.\space$^{4}$CUHK
$^{5}$Stanford \\$^{6}$NTU\space$^{7}$HKU\space$^{8}$University of Michigan\space$^{9}$SFU\space$^{10}$Google DeepMind\\
*equal contribution
}
\maketitle

\begin{center}
    \url{https://sherwinbahmani.github.io/tc4d}
\end{center}
\input{sec/0_abstract}    
\input{sec/1_intro}
\input{sec/2_related}
\input{sec/3_method}
\input{sec/4_experiments}
\input{sec/5_conclusion}

\bibliographystyle{splncs04}
\bibliography{egbib,ref}
\newpage
\appendix

\input{sec_supp/0_video_results}
\input{sec_supp/1_implementation}
\input{sec_supp/2_quantitative}
\input{sec_supp/3_trajectory}
\input{sec_supp/4_geometry}

%
%

\end{document}

%% file: sec/0_abstract.tex
\begin{abstract}
Recent techniques for text-to-4D generation synthesize dynamic 3D scenes using supervision from pre-trained text-to-video models.
However, existing representations, such as deformation models or time-dependent neural representations, are limited in the amount of motion they can generate---they cannot synthesize motion extending far beyond the bounding box used for volume rendering.
The lack of a more flexible motion model contributes to the gap in realism between 4D generation methods and recent, near-photorealistic video generation models.
Here, we propose TC4D: trajectory-conditioned text-to-4D generation, an approach that factors motion into global and local components.
We represent the global motion of a scene's bounding box using rigid transformation along a trajectory parameterized by a spline. 
We learn local deformations that conform to the global trajectory using supervision from a text-to-video model.
Our approach enables synthesis of scenes animated along arbitrary trajectories, compositional scene generation, and significant improvements to the realism and amount of generated motion, which we evaluate qualitatively and through a user study. Video results can be viewed on our website: \url{https://sherwinbahmani.github.io/tc4d}.
\end{abstract}

%% file: sec/1_intro.tex
\section{Introduction}
\label{sec:introduction}
Recent advances in video generation models~\cite{videoworldsimulators2024,wang2023modelscope,blattmann2023stable,singer2022make,ho2022imagen,blattmann2023align,wang2023videofactory,wu2023lamp} enable the synthesis of dynamic visual content with unprecedented fidelity.
These advances have prompted the rapid development of techniques for {\em 3D} video generation (see Fig.~\ref{fig:teaser}), which use supervision from video generation models to produce animated 3D scenes, conditioned on an input video~\cite{jiang2023consistent4d}, image~\cite{ren2023dreamgaussian4d}, or text prompt~\cite{singer2023text}.
With applications including visual effects, virtual reality, and industrial design, realistic and controllable generation of 4D content has broad potential impact.

\input{sec/0_teaser}
Although methods for 4D generation show immense promise, they are yet limited in the quality and realism of synthesized motion effects.
Specifically, these methods animate a volumetric scene representation via learned deformation~\cite{ling2023align,ren2023dreamgaussian4d,yin20234dgen} or appearance changes predicted by a neural representation~\cite{zhao2023animate124,zheng2023unified,jiang2023consistent4d,bahmani20234d} or generative model~\cite{pan2024fast}.  
Since the scene is confined to a particular region---e.g., a 3D bounding box containing the volumetric representation---the synthesized motion is highly local, and the magnitude of motion is typically much smaller than the scale of the scene itself. 
For example, existing techniques can synthesize characters that walk in place, gesticulate, or sway, but synthesized outputs cannot move around a scene or exhibit global motion effects in general.  
Overall, there is a stark contrast in realism between motion effects created with video generation models and existing techniques for 4D generation (see Fig.~\ref{fig:motivation_vid_4d}). 

To address this limitation, we consider the specific problem of text-to-4D generation with \textit{both local and global} motion, where global motions occur at scales larger than the bounding box of an object.
As existing architectures are incompatible with synthesizing global motion, we require a new motion representation that facilitates learning motion at coarse and fine scales.
Additionally, depicting global motion requires a large number of video frames, but current open-source video models typically output only a few tens of frames~\cite{blattmann2023stable}. 
Thus, we require a solution that decouples the temporal resolution and duration of generated 4D scenes from the output size constraints of the video generation model used for supervision.

Our solution to this problem uses trajectory conditioning for 4D generation (TC4D)---a form of coarse-scale motion guidance that enables a novel decomposition of global and local motion effects.
Specifically, we first use a text-to-3D model~\cite{bahmani20234d} to generate a static scene.
Then, we model global motion by animating the bounding box of an object using a rigid transformation along a 3D trajectory. 
We parameterize the trajectory using a spline specified by a user.
After fixing the global motion, we use supervision from a video diffusion model to learn local motion that is consistent with the global motion along the trajectory.

We model local motion along the trajectory by optimizing a time-varying deformation field based on a trajectory-aware version of video score distillation sampling (VSDS)~\cite{singer2023text}.
That is, we apply VSDS to arbitrarily long trajectories by stochastically sampling segments of the trajectory and supervising a deformation model that is used to animate the scene. 
We observe that the quality of synthesized motion effects using the deformation model is highly dependent on the diffusion time steps used with VSDS; sampling early timesteps introduces relatively no motion, but sampling later time steps leads to unrealistic, high-frequency jitter. 
To address this issue, we introduce an annealing procedure that gradually modifies the interval of sampled diffusion time steps during optimization to achieve temporally consistent motion. 
Altogether, trajectory conditioning enables coherent synthesis of local and global scene animation, dramatically improving the amount and realism of motion in generated 4D scenes.

In summary, we make the following contributions.
\begin{itemize}
    \item We introduce the task of 4D synthesis with global motion and propose a novel solution based on trajectory conditioning that decouples motion into global and local components.
    \item We demonstrate novel applications in text-to-4D generation with global motion, including 4D generation from arbitrary trajectories and compositional 4D scene synthesis.
    \item We conduct an extensive evaluation of the approach, including ablation and user studies, and we demonstrate state-of-the-art results for text-to-4D generation.
\end{itemize}

\input{fig/1_motivation_fig}

%% file: sec/0_teaser.tex
\begin{figure}[t!]
    \centering
    \includegraphics[width=\textwidth]{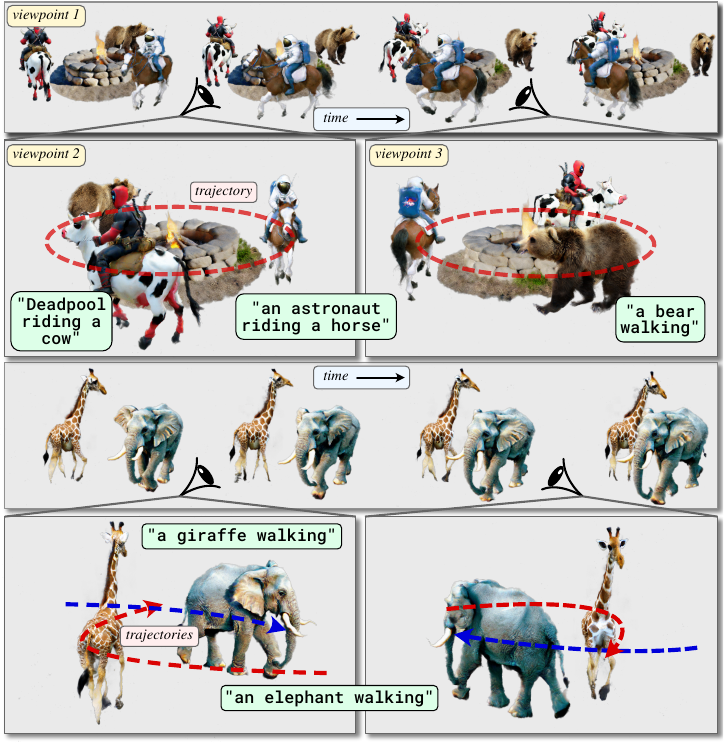}
    \captionof{figure}{\textbf{Scenes generated using trajectory-conditioned 4D generation (TC4D).} We show scenes consisting of multiple dynamic objects generated with text prompts and composited together. 
    The scene is shown for different viewpoints (panels) and at different time steps (horizontal dimension). 
Motion is synthesized by animating the scene bounding box along a provided trajectory using a rigid transformation, and we optimize for local deformations that are consistent with the trajectory using supervision from a video diffusion model. Overall, our approach improves the amount and realism of motion in generated 4D scenes.}
\label{fig:teaser}
\end{figure}

%% file: fig/1_motivation_fig.tex
\begin{figure}[t]
\centering
\includegraphics[width=\textwidth]{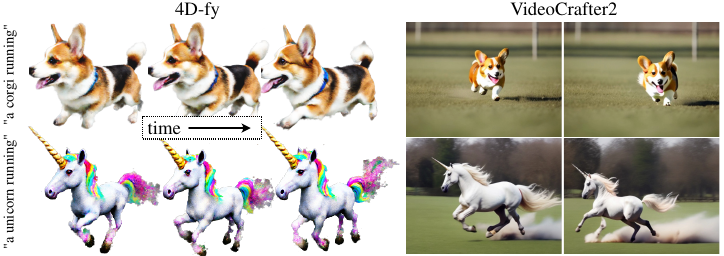}
\captionof{figure}{\textbf{Comparing text-to-video and text-to-4D generation.}
Existing methods for video generation (e.g., VideoCrafter2~\cite{chen2024videocrafter2}) create realistic objects with motion at both global and local scales. For example, a generated unicorn or corgi moves across the camera field of view (global motion) in concert with the animation of its legs and body (local motion). However, existing text-to-4D methods~\cite{bahmani20234d} only generate a limited amount of local motion and do not learn global motion in 3D space.}
\label{fig:motivation_vid_4d}
\end{figure}

%% file: sec/2_related.tex
\section{Related Work}
\label{sec:related}
Our work is most directly connected to text-to-video, text-to-3D, and text-to-4D generative models. For a comprehensive overview of recent advances in diffusion models and 4D generation, we refer readers to \cite{po2023state} and \cite{yunus2024recent}.

\subsubsection{Text-to-video generation.}
We build on recent advances in 2D image and video generative models, with methods for text-to-video generation being especially relevant to our approach~\cite{bie2023renaissance,po2023state}. While earlier methods for image and video generation favor GANs~\cite{skorokhodov2022stylegan, bahmani20223d}, recent work shifts towards using diffusion models for their superior performance in video generation~\cite{videoworldsimulators2024}. Still, most existing text-to-video models exhibit a significant quality gap compared to text-to-image models due to dataset and computational constraints.
Hybrid training approaches combining image and video datasets have been explored to overcome these limitations, employing techniques for enhancing spatial and temporal resolution through pixel-space and latent space upsampling~\cite{bain2021frozen,wang2023videofactory,xue2022advancing,ho2022imagen,guo2023animatediff,he2022latent,wang2023videocomposer,zhou2022magicvideo}. 
The emergence of transformer-based autoregressive models provides a leap forward, enabling the creation of videos of arbitrary lengths~\cite{vaswani2017attention,videoworldsimulators2024,ma2024latte,menapace2024snap}.
Recent efforts have also focused on fine-tuning text-to-image models for video data and decoupling content and motion generation for enhanced quality and flexibility~\cite{blattmann2023align,singer2022make,wu2023lamp,guo2023animatediff,blattmann2023stable}. Additionally, research aimed at increasing video generation controllability has shown promising results, allowing certain properties of an output video to be explicitly manipulated~\cite{wang2023motionctrl,wang2024boximator,shi2024motion,yang2024direct,bai2024uniedit}.

Despite recent advances, high-quality video synthesis remains challenging, with a notable quality disparity between open-source models and top-performing proprietary ones~\cite{zeroscope,wang2023modelscope}. Although the recent open-source method Stable Video Diffusion~\cite{blattmann2023stable} can synthesize high-quality videos from an input image, we use VideoCrafter2~\cite{chen2024videocrafter2} for our experiments due to its support for text conditioning.
\raggedbottom
\subsubsection{Text-to-3D generation.}
Early work on 3D generation leveraged the development of GAN algorithms to create realistic 3D objects of single categories, e.g., human or animal faces, cars, etc. These 3D GAN approaches \cite{devries2021unconstrained, chan2022efficient,or2022stylesdf,schwarz2022voxgraf, bahmani2023cc3d}, however, were limited in the class of scenes they could generate and thus were not suitable for text-based generation.
Initial attempts at data-driven, text-conditioned methods used CLIP-based supervision~\cite{radford2021learning}, which enabled synthesis or editing of 3D assets~\cite{chen2018text2shape,jain2022zero, sanghi2022clip,jetchev2021clipmatrix, gao2023textdeformer,wang2022clip}. These techniques evolved into the most recent approaches, which optimize a mesh or radiance field based on Score Distillation Sampling (SDS) and its variants using pre-trained diffusion models~\cite{poole2022dreamfusion, wang2023prolificdreamer, lin2022magic3d, chen2023fantasia3d, liang2023luciddreamer,wang2023score, li2024controllable, he2024gvgen, ye2024dreamreward, liu2023humangaussian, yu2023text, katzir2023noise, lee2023dreamflow, sun2023dreamcraft3d}. These SDS-based methods are improving, generating increasingly high-quality 3D assets.
The quality of generated 3D structures has been further improved by applying diffusion models that consider multiple viewpoints~\cite{lin2023consistent123, liu2023zero, shi2023mvdream, feng2024fdgaussian, liu2024isotropic3d, kim2023neuralfield, voleti2024sv3d, hollein2024viewdiff, szymanowicz2023splatter}.
These models alleviate the so-called ''Janus'' problem, where objects are generated with extra heads, limbs, etc.
Recent methods have shifted toward using diffusion or transformer models to lift an input 2D image into a 3D representation for novel-view synthesis~\cite{chan2023generative, tang2023make, gu2023nerfdiff, liu2023syncdreamer, yoo2023dreamsparse, tewari2024diffusion, qian2023magic123,long2023wonder3d, wan2023cad}. Other recent work~\cite{hong2023lrm,li2023instant3d,xu2023dmv3d,xu2024grm, zhang2024compress3d, han2024vfusion3d, jiang2024brightdreamer, xie2024latte3d, tang2024lgm, tochilkin2024triposr, qian2024atom} considers feed-forward 3D generation methods by first generating novel view images and converting them into 3D using transformer architectures.
Another line of work models compositional 3D scenes~\cite{po2023compositional, cohen2023set, bai2023componerf, chen2024comboverse, vilesov2023cg3d, epstein2024disentangled, zhou2024gala3d, gao2023graphdreamer, zhang2023scenewiz3d} by decomposing the scene into objects.
While these techniques for 3D generation produce impressive results, they do not support generating 4D scenes, i.e., their generated objects/scenes are static. 
We adapt MVDream~\cite{shi2023mvdream} and ProlificDreamer~\cite{wang2023prolificdreamer} to obtain an initial high-quality and static 3D model, which we animate along a trajectory using supervision from a video diffusion model.
\raggedbottom
\subsubsection{4D generation.}
Our research aligns closely with SDS-based 4D generation methodologies that are initialized with a 3D model and subsequently animated using text-to-video supervision. 
Within the last year, significant progress from initial results in 4D generation~\cite{singer2023text} has been made by a slew of recent methods~\cite{ren2023dreamgaussian4d, ling2023align,bahmani20234d,zheng2023unified, gao2024gaussianflow, yang2024beyond, jiang2023consistent4d, yang2024beyond}. 
These techniques have notably enhanced both geometry and appearance by integrating supervision from various text-to-image models. 
More recent endeavors~\cite{ren2023dreamgaussian4d,zhao2023animate124,yin20234dgen,pan2024fast, zheng2023unified, ling2023align, gao2024gaussianflow, zeng2024stag4d} have employed similar SDS-based strategies for image or video-conditioned 4D generation.
Still, all of these approaches yield 4D scenes that remain stationary at a single location and are limited to short and small motions, such as swaying or turning. 
Our objective, however, is to generate large-scale motions that more accurately mimic real-world movements.

%% file: sec/3_method.tex
\section{Trajectory-Conditioned Text-to-4D Generation}
\label{sec:method}

\input{fig/3_method_fig}

Our method generates a 4D scene given an input text prompt and trajectory. 
We assume that an initial static 3D scene can be generated by adapting existing approaches for text-to-3D generation~\cite{shi2023mvdream,wang2023prolificdreamer,bahmani20234d}, and we incorporate global and local motion effects using trajectory conditioning and trajectory-aware video score distillation sampling (VSDS).
Trajectory conditioning rigidly transforms the scene along a given path, and we parameterize local motions using a deformation model conditioned on the trajectory position. 
An overview of the method is shown in Fig.~\ref{fig:method}, and we describe each component as follows.

\subsection{4D Scene Representation}
We represent a 4D scene using a deformable 3D neural radiance field (NeRF)~\cite{mueller2022instant,li2022nerfacc,mildenhall2021nerf} parameterized by a neural representation $\mathcal{N}_\text{3D}$.
This representation models appearance and geometry within a unit-length bounding box $\mathbb{B} = \left[0,1\right]^3 $ as
\begin{equation}
    (\mathbf{c}, \sigma) = \mathcal{N}_\text{3D}(\mathbf{x}_c),\; \mathbf{x}_c \in \mathbb{B},
\end{equation}
where $\mathbf{c}$ and $\sigma$ are the radiance and density used for volume rendering~\cite{mildenhall2021nerf}.
The representation captures the canonical pose at \(t=t_0\) (without loss of generality we assume $t_0 = 0$), and it models motion of the scene at subsequent steps.
The 4D scene at any step \(0 < t \leq 1\) can be represented by warping the sampling rays with a time-varying deformation field \(\mathcal{F}(\cdot, t)\) to the canonical pose, \ie,
$
    \mathcal{N}_{\textrm{4D}}\left(\mathbf{x}, t\right) =  \mathcal{N}_{\textrm{3D}}\left(\mathcal{F}\left(\mathbf{x}, t\right)\right)$.

In this paper, we decompose the deformation $\mathcal{F}$ into a global and local term. 
The global motion is parameterized using a time-dependent rigid transformation $\mathcal{R}: (\mathbf{x}, t)\mapsto \mathbf{x}_\text{d}$ that applies rotation, translation and scaling at time $t$.
The local motion is given by a deformation field $\mathcal{D}: (\mathbf{x}_\text{d}, t) \mapsto \mathbf{x}_c $ that maps an input coordinate $\mathbf{x}_\text{d} \in \mathbb{B}$ in a deformed space to the coordinate in canonical (\ie, non-deformed) space $\mathbf{x}_\text{c} \in \mathbb{B}$.
Composing the global and local motions, the 4D scene can be represented as 
\begin{equation}
    \mathcal{N}_{\textrm{4D}}\left(\mathbf{x}, t\right) =  \mathcal{N}_{\textrm{3D}}\left(\mathcal{D}\circ\mathcal{R}\left(\mathbf{x}, t\right)\right).\label{eq:4drep}
\end{equation}
    
We describe the implementation of global and local motion operators in the next sections. 

\subsection{Modeling Global Motion with Trajectory Conditioning}
Our proposed motion decomposition enables using a straightforward model for global motion $\mathcal{R}$, by simply applying a rigid transformation to the bounding box $\mathbb{B}$ along a trajectory $\mathcal{T}$.
A holistic model of motion without this decomposition---\eg through deformation alone---would be intractable.
The scene bounding box would need to encompass the entire range of motion (inefficient in terms of computation and memory), and we would need to optimize long-range deformations, which is highly challenging.

In practice, the trajectory is parameterized as a cubic spline $\left\{\mathcal{T}\left(t\right)\vert t \in [0, 1]\right\}$ and $\mathcal{T}(0)=\mathbf{0}$ (\ie, the trajectory starts at the origin).
We assume the 3D control points are provided by a user.
This parametric curve defines the translation and the rotation of the bounding box $\mathbb{B}$, 
and the resulting global motion can be expressed as
\begin{equation}
    \mathcal{R}(\mathbf{x}, t) = \mathbf{R}(\mathbf{\hat{n}}, \mathcal{T}')\,\mathbf{x} + \mathcal{T}(t).
    \label{eq:rigid}
\end{equation}
Here, $\mathbf{R}$ is the rotation matrix that maps from the trajectory's tangent vector to the canonical bounding box orientation. 
Specifically, it maps the gradient of the trajectory $\mathcal{T}' = \mathrm{d} \mathcal{T} / \mathrm{d} t$ to the normal of the bounding box's ``front'' face $\mathbf{\hat{n}}$ while keeping the vertical component of the gradient perpendicular to the ground plane. 
Then, $\mathcal{T}\left(t\right)$ is the translation vector that maps back to the canonical bounding box location. 

\subsection{Modeling Local Motion with Trajectory-Aware VSDS}
We capture local motion by optimizing the deformation model with guidance from a trajectory-aware version of VSDS using a text-to-video diffusion model.
To regularize the generated deformation field we use a smoothness penalty, and we introduce a motion annealing technique, which empirically improves realism and temporal consistency.

\subsubsection{Video score distillation sampling (VSDS).}
For completeness, we briefly review the principle of VSDS, and refer the interested reader to Singer et al.~\cite{singer2023text} for a more detailed description.
Intuitively this procedure queries a video diffusion model to see how it adds structure to a video rendered from our representation.
Then, this signal is used to backpropagate gradients into the model to reconstruct the 4D scene. 

Specifically, a video sequence $\mathbf{v}$ is rendered from the 4D representation parameterized with $\theta$. 
Then, noise $\boldsymbol{\epsilon}$ is added to the video according to a diffusion process; the amount of noise is governed by the diffusion timestep $t_\text{d}$ such that $t_\text{d}=0$ yields a noise-free sequence and the maximum timestep ($t_\text{d}=1$) gives zero-mean Gaussian noise. 
The diffusion model, conditioned on a text embedding $\mathbf{y}$, predicts the noise added to the video ${\boldsymbol{\hat{\epsilon}}}(\mathbf{v}_{t_\text{d}}; t_\text{d}, \mathbf{y})$, and this is used to calculate the score distillation sampling gradient $\nabla_\theta \mathcal{L}_\text{4D}$ used to optimize the 4D representation~\cite{poole2022dreamfusion}.
\begin{equation}
    \nabla_\theta \mathcal{L}_\text{4D} = \mathbb{E}_{t_d, \boldsymbol{\epsilon}} \left[ \left( \boldsymbol{\hat{\epsilon}}(\mathbf{v}_{t_d}; t_d, \mathbf{y}) - \boldsymbol{\epsilon} \right) \frac{\partial\mathbf{v}}{\partial \theta}\right].
    \label{eq:vsds}
\end{equation}
Here, the expectation $\mathbb{E}$ is taken over all timesteps and noise realizations, and $ \frac{\partial\mathbf{v}}{\partial \theta}$ is the gradient of the generated video \wrt the 4D representation parameters $\theta$. 
In our case, we freeze the parameters of $\mathcal{N}_\text{3D}$ and optimize the parameters of the deformation model $\mathcal{D}$ to learn local motion. 

\subsubsection{Incorporating the trajectory.}
One challenge with using VSDS to generate dynamic objects along a trajectory is that the pre-trained video diffusion models generate a finite number of $N$ frames and thus can only represent limited range of motion. 
Naively sampling $N$ timesteps across a long trajectory would require the diffusion model to represent long-range motion, different from its training data.

Instead, we divide the trajectory into segments of length $\Delta t$ that roughly align with the intrinsic motion scale of the video diffusion model---that is, the range of motion typically exhibited in generated video samples. 
Assuming the intrinsic motion scale $M_\mathbf{v}$ of the video model is known, and given the total length of the trajectory $L$, we approximate the segment length as 
\begin{equation}
\Delta t = \max\left( \frac{M_\mathbf{v}}{L}, 1\right).    \label{eq:delta_t}
\end{equation}
In practice, we determine $M_\mathbf{v}$ based on an empirical evaluation across multiple scene categories (see supplement for additional details).

Our trajectory-aware VSDS thus consists of three steps:
(1) we sample random start times $t_0 \in [0, 1-\Delta t]$ along the trajectory, and evenly sample $N$ frame times as $t_n = t_0 + n \Delta t/(N-1)$ for $0 \leq n \leq N - 1$; (2) we compute the corresponding scene representation from $\mathcal{N}_\text{4D}$ using Eq.~\ref{eq:4drep};  (3) we render videos and optimize the deformation model using Eq.~\ref{eq:vsds}. 
We summarize the entire optimization procedure Algorithm~\ref{alg:tc4d}. 

\begin{algorithm}[t]
\caption{TC4D}\label{alg:cap}
\label{alg:tc4d}
\begin{flushleft}
\vspace{0.5em}
\textbf{Require:} \\
\hspace*{\algorithmicindent} $\mathcal{N}_\text{3D}$ \Comment{optimized 3D neural representation}\\ 
\hspace*{\algorithmicindent} $\mathcal{T}$ \Comment{trajectory parameterized with a spline}\\
\hspace*{\algorithmicindent} $\mathcal{D}$ \Comment{initial local deformation model parameterized with $\theta$}\\
\textbf{Output:} \\
\hspace*{\algorithmicindent} $\mathcal{D}^*$ \Comment{optimized local deformation model}
\end{flushleft} 
\hrule\vspace{0.5em}
\begin{algorithmic}[1]    
    \State determine $\Delta t$
    \State sample $t_0$ and trajectory steps 
    $t_n,\; 0 \leq n \leq N-1$
    \State sample points along rendering rays $\mathbf{x}$
    \State apply rigid transform and local deformation\ (Eq.~\ref{eq:4drep})

    $\mathbf{x}_\text{c} = \mathcal{D}\circ\mathcal{R}(\mathbf{x}, t_n)$
    \State volume render video from $\mathcal{N}_\text{3D}(\mathbf{x}_\text{c})$ and render deformation video from $\mathcal{D}(\mathbf{x}, t)$
    \State calculate $\nabla_\theta \mathcal{L}_\text{4D}$ and $\nabla_\theta \mathcal{L}_{\text{smooth}}$ from Eq.~\ref{eq:vsds} and Eq.~\ref{eq:total_variation}
    \State update $\mathcal{D}$ 
    \State repeat 1-7
\end{algorithmic}
\end{algorithm}

\subsection{Implementation details}
\label{sec:method_impl}
Following prior work~\cite{zheng2023unified, bahmani20234d}, we use hash-encoded feature grids to encode the 3D NeRF $\mathcal{N}_\text{3D}$ and the local deformation field \(\mathcal{D}\). $\mathcal{N}_\text{3D}$, is optimized using the hybrid SDS proposed by Bahmani~\etal~\cite{bahmani20234d}, which incorporates supervision from a combination of pre-trained image diffusion models~\cite{shi2023mvdream,wang2023prolificdreamer} to generate a static 3D scene with high-quality appearance and 3D structure.

Once trained, we freeze $\mathcal{N}_\text{3D}$ and optimize the local deformation field $\mathcal{D}$ using the trajectory-aware VSDS gradient and an additional smoothness loss.
For ease of optimization, we let $\mathcal{D}$ represent the offset $\mathbf{d} = \mathbf{x}_\text{d}-\mathbf{x}_\text{c}$ and initialize it to a small value.
Following Zheng et al.~\cite{zheng2023unified}, we apply volume rendering to create videos of the 3D displacement and compute the penalty as 
\begin{equation}
    \mathcal{L}_{\text{smooth}} = \lambda\sum\limits_{\mathbf{p}, n} \lVert \mathbf{d}_x \rVert_2^2 + \lVert \mathbf{d}_y \rVert_2^2 + \lVert \mathbf{d}_t \rVert_2^2,
    \label{eq:total_variation}
\end{equation}
where the summation is over all pixels $\mathbf{p}$, video frames $n$, $\lambda$ is a hyperparameter, and $\mathbf{d}_{x/y/t}$ represents a finite-difference approximation to spatial and temporal derivatives in the 3D displacement video sequence.

Additionally, we find that VSDS gradients for diffusion timesteps where $t_\text{d} \approx 1$ introduce jittery, high-frequency motion into the deformation field. 
We address this issue by annealing the sampling interval of the diffusion timesteps during the course of optimization, which improves the temporal coherence of local motion.
At the start of training we compute $\nabla\mathcal{L}_\text{4D}$ by uniformly sampling $t_d \in [0.02, 0.98]$. 
This interval is linearly decayed during training to $t_d \in [0.02, 0.5]$.

The model is trained using the Adam optimizer~\cite{kingma2020method} for 10000 iterations, which we find is sufficient to ensure robust and repeatable convergence.

%% file: fig/3_method_fig.tex
\begin{figure}[t]
\centering
\includegraphics[width=\textwidth]{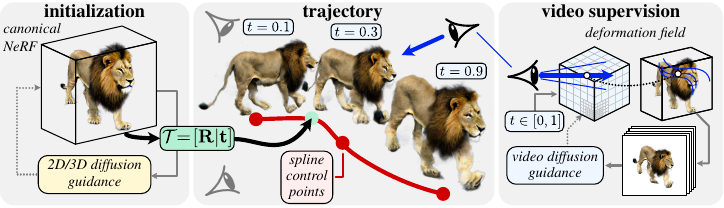}
\captionof{figure}{\textbf{Overview of TC4D.} Our method takes as input a pre-trained 3D scene generated using supervision from a 2D and/or 3D diffusion model. We animate the scene through a decomposition of motion at global and local scales. Motion at the global scale is incorporated via rigid transformation of the bounding box containing the object representation along a given spline trajectory $\mathcal{T}$ at steps $t$. We align local motion to the trajectory by optimizing a separate deformation model that warps the underlying volumetric representation based on supervision from a text-to-video model. The output is an animated 3D scene with motion that is more realistic and greater in magnitude than previous techniques.}
\label{fig:method}
\end{figure}

%% file: sec/4_experiments.tex
\section{Experiments}
\label{sec:experiments}
Motion quality is extremely difficult to visualize. 
In this manuscript, we focus on presenting a detailed analysis of our quantitative evaluation and include only the most representative visual examples. 
We highly encourage the readers to check the videos in the supplemental material for a conclusive quality assessment.

\subsection{Metrics}
\label{sec:metrics}

Since there is no automated metric for 4D generation, we evaluate TC4D with a user study. We refer to Sec.~3 of the appendix for additional CLIP score evaluations.
The method is compared quantitatively and qualitatively to a modified version of 4D-fy~\cite{bahmani20234d}.  
That is, we animate the output of 4D-fy by rigidly transforming it to follow the same trajectory as used for TC4D. We further compare to the concurrent work DreamGaussian4D~\cite{ren2023dreamgaussian4d} in Sec.~3 of the appendix.
In total we provide comparisons using 33 scenes generated from different text prompts, and we conduct an extensive ablation study on a subset of these text prompts.
Our evaluation uses a similar number of text prompts as previous work (e.g., 28~\cite{singer2023text}, or 17~\cite{bahmani20234d}).

\input{fig/4_main_fig}

We recruit a total of 20 human evaluators and show the evaluators two video clips generated using the same text prompt with TC4D and 4D-fy. 
We ask the evaluators to specify their overall preference and their preference in terms of appearance quality (AQ), 3D structure quality (SQ), motion quality (MQ), motion amount (MA), text alignment (TA) (see the supplement for the detailed protocol for the study). The results of the user study are shown in Table~\ref{tab:results} and we include further details in the supplement.
Each row of the table reports the percentage of evaluations that ranked T4CD over 4D-fy or the ablated approach, \ie, a percentage above 50\% indicates T4CD is preferred. 
Statistically significant results ($p<0.05$) are indicated based on a $\chi^2$ test.

\input{tab/4_main_tab}

\subsection{Results}
\label{sec:results}
We visualize the generated 4D scenes by rendering and compositing images along the trajectory from multiple camera viewpoints in Fig.~\ref{fig:main_sota}.
TC4D shows significantly more motion and also more realistic motion than 4D-fy---this is especially noticeable for scenes in which TC4D synthesizes walking animations that are coherent with the trajectory, but 4D-fy produces only a small amount of relatively incoherent motion.
In the user study, participants indicated a statistically significant preference for results from TC4D for all metrics except structure quality. 
In terms of overall preference, 85\% of comparisons were in favor of T4CD.
We refer the reader to the video clips included in the supplement, which provide a clear visualization of the advantages of the proposed approach. In addition to these results based on user-defined trajectories, we show results for automated end-to-end generation using large language models in Sec.~4 of the appendix. Moreover, we provide geometry visualizations in Sec.~5 of the appendix

\subsection{Ablation Study}
\label{sec:ablations}

We assess the impact of various design choices for TC4D through an extensive ablation study.
The results are evaluated in the user study (Table~\ref{tab:results}), and we show qualitative results for the ablated methods on multiple scenes in video clips included in the supplement. To allow for fairer comparison between the methods, we initialize each method with the same static model for a given prompt. This generally led to similar appearance and 3D structure across all methods, which is why many of the quantitative results for appearance quality and 3D structure quality in Table~\ref{tab:results} do not indicate a statistically significant preference.
We discuss each of the ablations in the following.

\paragraph{Local deformation.}
Removing the local deformation model reduces the method to rigid transformation of a static pre-trained 3D scene along a spline (i.e., with $\mathcal{N}_\text{4D}(\mathbf{x}, t) = \mathcal{N}_\text{3D}(\mathcal{R}(\mathbf{x}, t))$). 
This ablated method produces unrealistic and completely rigid motion.
It is indicated in Table~\ref{tab:results} as ``w/o local deformation''. 
The results show that the participants perceive less and lower quality motion (see MA and MQ columns respectively), and they overwhelmingly prefer including the deformation motion component.

\paragraph{Trajectory (training and rendering).}
Alternatively, entirely removing the trajectory reduces the method to the conventional text-to-4D generation setting with deformation-based motion.
That is, we have $\mathcal{N}_\text{4D} = \mathcal{N}_\text{3D}(\mathcal{D}(\mathbf{x}, t))$.
This experiment is indicated in Table~\ref{tab:results} as ``w/o trajectory''. 
While the method still learns to incorporate some motion without the trajectory, we find that the trajectory is key to producing more complex animations such as walking.

\paragraph{Trajectory (training).}
Similar to the 4D-fy baseline, we can omit the trajectory at the training stage and animate the scene via rigid transformation after training. 
This experiment is indicated in Table~\ref{tab:results} as ``w/o traj. training''. 
In this experiment, since the synthesized motion is agnostic to the trajectory, the resulting animations are less coherent than when incorporating the trajectory into the training procedure.
Specifically, only a smaller range of motion is visible. 
User study participants prefer TC4D over this ablation by a significant margin.

\paragraph{Trajectory-aware VSDS.}
We use trajectory-aware VSDS to ensure that all regions of the trajectory receive adequate supervision from the text-to-video model.
Specifically, this procedure involves breaking the trajectory down into segments, and then sampling one segment for which we render the video clips used to compute the VSDS gradients.
Then the VSDS gradients are computed using the resulting video.
Overall, we find this approach results in too coarse a sampling of the trajectory and fails to synthesize smooth motion.
In Table~\ref{tab:results}, this method is indicated as ``w/o traj.-aware VSDS''.
\paragraph{Global transformations.}
To demonstrate that VSDS is trajectory dependent, we conduct an ablation where we remove global transformations during training, but still sample the time dimension in a segmented way as in trajectory-aware VSDS. The local deformations in this approach are significantly reduced.

\paragraph{Smoothness penalty.}
We assess the impact of removing the smoothness penalty from the deformation field (Eq.~\ref{eq:total_variation}). 
This regularization term, proposed by Zheng et al.~\cite{zheng2023unified}, helps to dampen high-frequency displacements in the deformation field and leads to smoother motion. 
As the row ``w/o smooth penalty'' shows in Table~\ref{tab:results}, the responses from user study participants indicate that removing this penalty has only a modest negative effect.

\paragraph{Diffusion timestep annealing.}
Scheduling the sampling of the diffusion timesteps during training with VSDS provides another form of motion regularization in our method.
Specifically, we reduce the sampling of large diffusion timesteps as we train, and we find that this produces smoother, more realistic motion.
As we show in the last row of Table~\ref{sec:results}, ``w/o timestep annealing'',
removing this procedure entirely (and keeping the smoothness penalty) negatively impacts the users' preference. 
In the supplement, we show that this method produces jittery animations and noticeable high-frequency deformation.

\paragraph{Effect of scale.}
Finally, we evaluate the effect of combining scaling with the rigid transformation used to define the trajectory.
That is, we modify Eq.~\ref{eq:rigid} to  incorporate a time-dependent term $\mathcal{S}(t) \in \mathbb{R}_+^3$, which scales the bounding box along each axis:
\begin{equation}
   \mathcal{R}_\mathcal{S}(\mathbf{x}, t) = \mathcal{S}(t) \odot \mathbf{R}(\mathbf{\hat{n}}, \mathcal{T}')\,\mathbf{x} + \mathcal{T}(t),
\label{eq:rigid_scale}
\end{equation}
where $\odot$ indicates Hadamard product.

We demonstrate this approach on a scene consisting of a column of fire that is generated with the text prompt, ``a flame getting larger'' (Fig.~\ref{fig:flame}).
Generating this scene without the trajectory and without bounding box scaling results in a flame that has little variation in size.
Alternatively, we animate the flame along an upwards trajectory, but this approach simply translates the generated scene rather than changing the size.
Combining translation and scaling in the vertical dimension allows us to keep the center of the bounding box at a consistent location, and with this technique, the generated flame appears to grow in size.
\input{fig/4_flame_fig}

%% file: fig/4_main_fig.tex
\begin{figure}[h!]
\centering
\includegraphics[width=\textwidth]{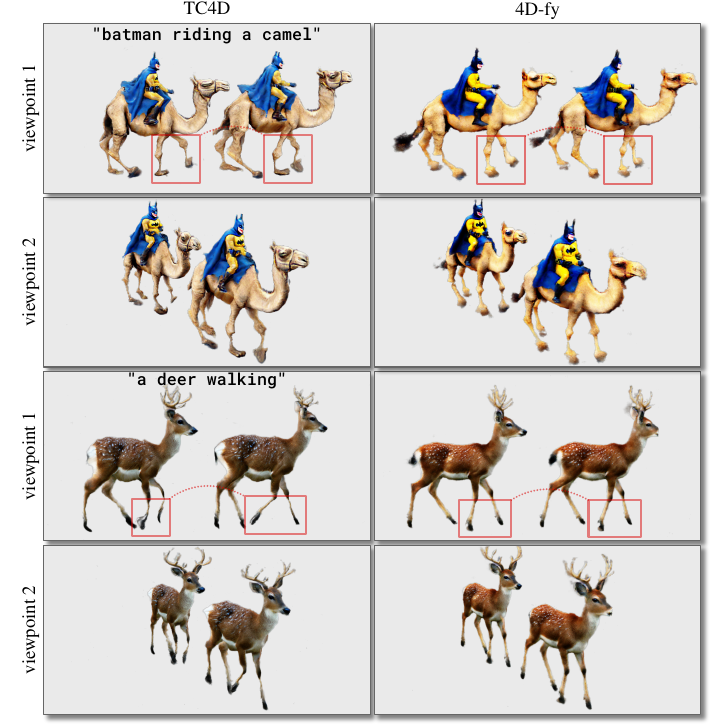}
\captionof{figure}{\textbf{Comparison of TC4D and 4D-fy.}
 We show two generated scenes of ``batman riding a camel'' and ``a deer walking''. Each panel contains images rendered from a single viewpoint from two steps along the same trajectory. While 4D-fy produces mostly static results (which we rigidly transform along the trajectory), TC4D learns coherent motion and produces a walking animation. Please also refer to the video results in the supplement.}
\label{fig:main_sota}
\end{figure}

%% file: tab/4_main_tab.tex
\begin{table}[!t]
    \caption{\textbf{Quantitative results.} We compare TC4D to the output of 4D-fy animated along the trajectory using a rigid transformation. The methods are evaluated in a user study in which participants indicate their preference (or no preference) based on appearance quality (AQ), 3D structure quality (SQ), motion quality (MQ), motion amount (MA), text alignment (TA), and overall preference (Overall). We also conduct ablations using a subset of the text prompts. The percentages indicate preference for TC4D vs. the alternative method (in each row), \ie, >50\% indicates ours is preferred. All results are statistically significant ($p < 0.05$) except those indicated with an asterisk.}
    \label{tab:results}
    \begin{center}
    \begin{tabular}{l|ccccc|c}
        \toprule
         & \multicolumn{6}{c}{\textit{Human Preference (Ours vs. Method)}}\\
        \textit{Method} & AQ & SQ & MQ & MA & TA & Overall \\\midrule
        4D-fy~\cite{bahmani20234d}  & 55\% & 52*\% & 89\% & 92\% & 81\% & 85\%\\
        \midrule
        \textit{Ablation Study} & \multicolumn{5}{c}{}  \\\midrule
        w/o local deformation  &59\%* & 57\%* & 98\% & 98\% & 86\% & 97\% \\
        w/o trajectory  & 56\%* & 61\% & 80\% & 85\% & 79\% & 84\% \\
        w/o traj. training  & 61\% & 56\%* & 87\% & 92\% & 79\% & 89\% \\
        w/o traj.-aware VSDS  & 62\% & 59\%* & 84\% & 84\% & 74\% & 85\% \\
        w/o global transformations  & 65\% & 58\%* & 94\% & 94\% & 85\% & 97\% \\
        w/o smooth. penalty  & 58\%* & 60\%* & 73\% & 69\% & 63\% & 74\% \\
        w/o timestep annealing  & 63\% & 65\% & 84\% & 83\% & 73\% & 84\% \\
        \bottomrule
    \end{tabular}
    \end{center}
    \vskip -0.2in
\end{table}

%% file: fig/4_flame_fig.tex
\begin{figure}[t!]
    \vspace{0em}
\centering
\floatbox[{\capbeside\thisfloatsetup{capbesideposition={right,top},capbesidewidth=5.5cm}}]{figure}[\FBwidth]
{\caption{\textbf{Trajectory with scale.} We demonstrate adding a scale term to the trajectory for a scene generated with the text prompt ``a flame getting larger''. We compare generating the scene without a trajectory or scale (top), with an upwards trajectory only (middle), and with both a trajectory and scale (bottom), the last of which yields a convincing result.}\label{fig:flame}}
{\includegraphics[]{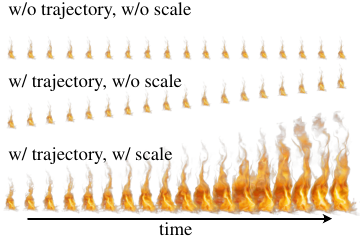}}
\end{figure}

%% file: sec/5_conclusion.tex
\section{Conclusion}
\label{sec:conclusion}

Overall our work takes important steps in making motion synthesis for text-to-4D models more realistic and expressive. Our framework enables scene-scale motion of entities within compositional 4D scenes with trajectory conditioning, achieving improvement over prior work that shows motion at fixed locations. 

The main limitations of our results are imperfect leg motion, foot sliding, or asymmetric and not fully cyclic motion. 
The motion imperfections stem from the quality of publicly available pre-trained text-to-video models used for VSDS.
Also, VideoCrafter2 outputs 16-frame videos that may be too short to capture fully cyclic motion (e.g., a complete walking cycle), contributing to this limitation.
We expect that TC4D will benefit from rapid improvements in 2D video generation, and we open source our code to spur follow-on work.

The proposed factorized motion model opens several avenues for future work.
In particular, we envision extensions to generate scenes with multiple interacting objects, adding motion to existing in-the-wild 3D assets, and synthesis of motion across unbounded, large-scale scenes.
Finally, designing automated text-to-4D metrics by temporally extending text-to-3D metrics~\cite{wu2024gpt} is an important direction for future work.

\paragraph{Ethics statement.}
Our approach can automatically generate realistic, animated 3D scenes; such techniques can be misused to promulgate realistic fake content for the purposes of misinformation. 
We condemn misuse of generative models in this fashion and emphasize the importance of research to thwart such efforts (see, e.g., Masood et al.~\cite{masood2023deepfakes} for an extended discussion of this topic).

\section*{Acknowledgements}

This work was supported by the Natural Sciences and Engineering Research Council of Canada (NSERC) Discovery Grant program, the Digital Research Alliance of Canada, and by the Advanced Research Computing at Simon Fraser University. It was also supported by Snap Inc. and a Swiss Postdoc Mobility fellowship.

%% file: sec_supp/0_video_results.tex
\section{Video Results}
\label{sec:supp_vid_res}

We include an extensive set of 4D generation results in video format in the supplementary website \url{https://sherwinbahmani.github.io/tc4d}, best viewed with \textbf{Google Chrome}.
There, we showcase compositional 4D scenes, a diverse set of trajectory-conditioned 4D results and comparisons to 4D-fy~\cite{bahmani20234d}.

%% file: sec_supp/1_implementation.tex
\section{Implementation Details}
\label{sec:supp_impl}

\paragraph{SDS guidance.}
We implement our approach based on the threestudio framework~\cite{threestudio2023}, which includes implementations of MVDream~\cite{shi2023mvdream}~for 3D-aware text-to-image diffusion and score distillation sampling (SDS), ProlificDreamer~\cite{wang2023prolificdreamer} with Stable Diffusion~\cite{rombach2022high}~(text-to-image diffusion and VSD), and we implement video SDS using VideoCrafter2~\cite{chen2024videocrafter2}.

\paragraph{Hyperparameter values.}
We initialize the 4D neural representation following \cite{poole2022dreamfusion, lin2022magic3d} and add an offset to the density predicted by the network in the center of the scene to promote object-centric reconstructions in the static stage.
For the deformation network, we use a 4D hash grid with 8 levels, 2 features per level, and a base resolution of 4. 
We decode the dynamic hash grid features into a 3D deformation vector with a small two layer multi-layer perceptron (MLP) with 64 hidden features.
We set the learning rates to 0.01 for the static hash map, 0.001 for the deformation hash map, and 0.0001 for the deformation MLP.
The number of training iterations for the two static stages of hybrid SDS~\cite{bahmani20234d} as well as the dynamic stage are set to 10000, 10000, and 10000, respectively.
We set the probability for hybrid SDS in the static stage to 0.5, following 4D-fy~\cite{bahmani20234d}.

The intrinsic motion scale $M_\mathbf{v}$ is set to 0.3 for all text prompts. Increasing $M_\mathbf{v}$ leads to larger jumps between each timestep and a tendency towards low-frequency motion; smaller values of $M_\mathbf{v}$ lead to high motion variation between nearby samples on the trajectory and high-frequency motion. 
Overall, this hyperparameter depends on the specific text prompt, and per prompt tuning could yield improved results. 
For simplicity and a fair comparison to 4D-fy, which does not tune any hyperparameters per text prompt, we set $M_\mathbf{v}$ to be the same across all text prompts. 

\paragraph{Rendering.}
Each of the pretrained diffusion models has a different native resolution, so we render images from network accordingly.
We render four images from different camera positions for the 3D-aware SDS update at the native (256$\times$256 pixel) resolution of the 3D-aware text-to-image model.
The VSD update is computed by rendering a 256$\times$256 image and bilinearly upsampling the image to the native resolution of Stable Diffusion (512$\times$512).
Finally, the video SDS update is computed by rendering 16 video frames at 128$\times$80 resolution and upsampling to the native 512$\times$320 resolution of VideoCrafter2. 
To render a background for the image, we optimize a second small MLP that takes in a ray direction and returns a background color. At inference time, we render all videos on the same grey background.

\paragraph{Camera pose sampling.}
For training, we sample one random view per VSDS update shared across all trajectory timesteps. For each iteration, we randomly sample a camera on a sphere with radius in [1.5, 2.0], field of view in [15$^{\circ}$, 60$^{\circ}$], elevation in [0$^{\circ}$, 30$^{\circ}$], and azimuth in [-180$^{\circ}$, 180$^{\circ}$]. 
For inference, we render videos with radius 1.75, field of view 40$^{\circ}$, elevation 15$^{\circ}$, and equidistant azimuth samples in [-180$^{\circ}$, 180$^{\circ}$]. The radius and field of view are selected to ensure that objects largely stay within the image frames.

\paragraph{Computation.}
We optimized the model on a single NVIDIA A100 GPU. The entire procedure requires roughly 80 GB of VRAM and the three optimization stages require approximately 1, 2, and 7 hours of compute, respectively. 
The first two optimization stages follow the static optimization procedure of 4D-fy, and the third optimization stage involves training our deformation model.
In comparison to the baseline 4D-fy, which takes 23 hours in total, our third stage training takes significantly less time to converge while demonstrating more motion.

\paragraph{4D-fy baseline.}
While 4D-fy~\cite{bahmani20234d} is an object-centric text-to-4D method, we extend it to a trajectory-conditioned generation method for a fair comparison. Concretely, we use identical trajectories as for our methods as input and apply the trajectory post-training as a naive approach to combining text-to-4D generation with a trajectory. We use the same video diffusion model VideoCrafter2~\cite{chen2024videocrafter2} as guidance for TC4D and 4D-fy for a fair comparison. 

%% file: sec_supp/2_quantitative.tex
\section{Supplemental Experiments}
\label{sec:supp_exp}

\paragraph{Comparison with DreamGaussian4D.}

\begin{figure}
    \centering
    \includegraphics[]{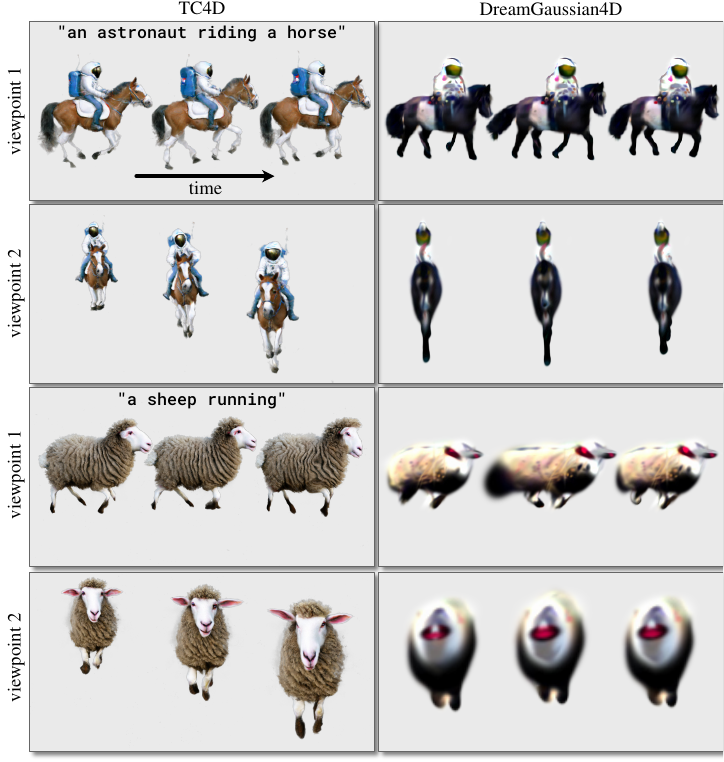}
    \caption{\textbf{Comparison to DreamGaussian4D.} We compare 4D scenes generated with TC4D and DreamGaussian4D~\cite{ren2023dreamgaussian4d}. The proposed approach generates more realistic scenes with greater amounts of motion (see supplemental video).}
    \label{fig:supp-dreamgaussian4d}
\end{figure}

We further compare our method with the concurrent work DreamGaussian4D~\cite{ren2023dreamgaussian4d}. Similar to the 4D-fy baseline, we naively extend DreamGaussian4D by animating the generated models post-hoc using same trajectories as TC4D for fair comparison. 
Note that DreamGaussian4D is an image-to-4D method, hence we generate the input image with StableDiffusion 2.1 \cite{rombach2022high}. We notice that the 3D structure of DreamGaussian4D is sensitive to the input image being perfectly aligned in a front facing view. 
While re-running StableDiffusion with multiple seeds can minimize the issue, this does not provide a fully automated approach. 
We believe that some sort of initial pose estimation as part of the DreamGaussian4D pipeline could further improve the results; however this is out of the scope of our comparison as it requires architectural adjustments to the DreamGaussian4D baseline. 
In contrast, our method is robust to varying random seeds as we do not rely on front facing input images. We show a visual comparison in Fig.~\ref{fig:supp-dreamgaussian4d} and quantitative results in Tab.~\ref{tab:results_supp}. We additionally ran a separate user study with 17 participants who each evaluated 33 comparisons between our method and the DreamGaussian4D baseline and obtained results overwhelmingly in our favor, which we show in Tab.~\ref{tab:full_results}.

\paragraph{CLIP Score.} In addition to the user study results, we evaluate all methods using CLIP score following the protocol in 4D-fy \cite{bahmani20234d}. 
CLIP Score~\cite{park2021benchmark} evaluates the correlation between a text prompt and an image. Specifically, this corresponds to the cosine similarity between textual CLIP \cite{radford2021learning} embedding and visual CLIP \cite{radford2021learning} embedding. 
For all text prompts, we render a video of the generated 4D scene following the camera trajectory used 4D-fy (i.e., the camera moves in azimuth at a fixed elevation angle).
All rendered video frames are scored using CLIP ViT-B/32, and we report the average CLIP score over all frames and text prompts.

We show CLIP scores in Tab.~\ref{tab:results_supp}.
Note that except for DreamGaussian4D, which generally shows low per-frame quality, the scores are very similar across methods and ablations. 
The CLIP model only takes a single frame as input, and so it cannot be expected to assess motion quality. 
Since the main difference between our method and the baselines relates to motion along the trajectory, we find that the CLIP score is a relatively weak evaluation metric.
Nevertheless, for the sake of completeness, we include CLIP scores below. 
We note that the user study and supplementary video results clearly show the motion benefits of our method. 
Developing automated video metrics for text-to-4D generation is an important direction for future work.

\begin{table}[!t]
    \caption{\textbf{CLIP Score results.} Note that the ablations were evaluated with a subset of the prompts to save computational costs, hence the ablation CLIP score of our method differs from the main CLIP score.}
    \label{tab:results_supp}
    \begin{center}
    \begin{tabular}{lc}
        \toprule
        \textit{Method} & CLIP \\\midrule
        Ours  & 32.01 \\
        4D-fy  & 31.90 \\
        DreamGaussian4D  & 28.70 \\
        \midrule
        \textit{Ablation Study} & \multicolumn{1}{c}{}  \\\midrule
        Ours  & 32.69\\
        w/o local deformation  & 32.11 \\
        w/o trajectory  & 32.16 \\
        w/o traj. training  & 32.16 \\
        w/o traj.-aware VSDS  & 32.62 \\
        w/o smooth. penalty  & 32.68\\
        w/o timestep annealing  & 32.62 \\
        \bottomrule
    \end{tabular}
    \end{center}
    \vskip -0.2in
\end{table}

\paragraph{User Study.} We give additional details of our user study. Following the protocol in 4D-fy \cite{bahmani20234d}, users were prompted with 63 pairs of videos. Each pair contains a result generated from our method and from a comparison method, with the same prompt used to generate both. The evaluator is then asked to select their preference between each video, or to indicate equal preference. 
In the main paper, we aggregate selections for ours and ties, where ties count equally towards ours and theirs. 
In Tab.~\ref{tab:full_results}, we show the full breakdown between ours, ties, and theirs. As can be seen, the majority of evaluations for appearance quality and 3D structure quality are ties, while the majority of evaluations for motion quality and motion amount prefer our the proposed method.

\begin{table}[!t]
    \caption{\textbf{Full user study results.} We compare TC4D to modified versions of 4D-fy and DreamGaussian4D. The methods are evaluated in a user study in which participants indicate their preference (or no preference) based on appearance quality (AQ), 3D structure quality (SQ), motion quality (MQ), motion amount (MA), text alignment (TA), and overall preference (Overall). We also conduct ablations using a subset of the text prompts. The percentages indicate user selections for ours/ties/theirs.}
    \label{tab:full_results}
    \begin{center}
    \resizebox{\columnwidth}{!}{
    \begin{tabular}{l|ccc}
        \toprule
         & \multicolumn{3}{c}{\textit{Human Preference (Ours vs. Method)}}\\
        \textit{Method} & AQ & SQ & MQ \\\midrule
        4D-fy  & 31\%/48\%/21\%  & 28\%/49\%/23\% & 82\%/11\%/7\% \\
        DreamGaussian4D & 99.8\%/0.0\%/0.2\% & 99.8\%/0.0\%/0.2\% & 99.6\%/0.2\%/0.2\% \\
        \midrule
        \textit{Ablation Study} & \multicolumn{3}{c}{}  \\\midrule
        w/o local deformation  & 29\%/58\%/13\%  & 29\%/56\%/15\% & 94\%/6\%/0\% \\
        w/o trajectory  & 32\%/52\%/16\%  & 34\%/55\%/11\% & 72\%/20\%/8\% \\
        w/o traj. training  & 25\%/68\%/7\%  & 22\%/66\%/12\% & 78\%/16\%/6\% \\
        w/o traj.-aware VSDS  & 26\%/69\%/5\%  & 23\%/70\%/7\% & 72\%/23\%/5\% \\
        w/o global transformations  & 34\%/62\%/4\%  & 24\%/68\%/8\% & 94\%/5\%/1\% \\
        w/o smooth. penalty  & 23\%/69\%/8\%  & 24\%/69\%/7\% & 48\%/44\%/8\% \\
        w/o timestep annealing  & 29\%/63\%/8\%  & 28\%/69\%/3\% & 64\%/34\%/2\% \\
        \bottomrule
    \end{tabular}}
    \end{center}
    \begin{center}
    \resizebox{\columnwidth}{!}{
    \begin{tabular}{l|cc|c}
        \toprule
         & \multicolumn{3}{c}{\textit{Human Preference (Ours vs. Method)}}\\
        \textit{Method} & MA & TA & Overall \\\midrule
        4D-fy  & 88\%/6\%/6\% & 64\%/33\%/3\% & 78\%/13\%/9\% \\
        DreamGaussian4D & 99.3\%/0.5\%/0.2\% & 98.5\%/1.3\%/0.2\% & 99.8\%/0.0\%/0.2\% \\
        \midrule
        \textit{Ablation Study} & \multicolumn{2}{c}{}  \\\midrule
        w/o local deformation  & 96\%/3\%/1\% & 75\%/24\%/1\% & 93\%/6\%/1\% \\
        w/o trajectory  & 78\%/15\%/7\% & 66\%/30\%/4\% & 78\%/14\%/8\% \\
        w/o traj. training  & 87\%/9\%/4\% & 60\%/38\%/2\% & 82\%/13\%/5\% \\
        w/o traj.-aware VSDS  & 72\%/22\%/6\% & 48\%/49\%/3\% & 73\%/22\%/5\% \\
        w/o global transformations  & 93\%/2\%/5\%  & 71\%/29\%/0\% & 94\%/6\%/0\% \\
        w/o smooth. penalty  & 48\%/38\%/14\% & 30\%/64\%/6\% & 54\%/34\%/12\% \\
        w/o timestep annealing  & 65\%/32\%/3\% & 44\%/54\%/2\% & 68\%/28\%/4\% \\
        \bottomrule
    \end{tabular}}
    \end{center}
    \vskip -0.2in
\end{table}

%% file: sec_supp/3_trajectory.tex
\section{Automated Generation}
\label{sec:supp_auto}

The trajectory control points used to generate results shown in the main text are set manually---this facilitates generating a diverse set of results. 
Moreover, using a sparse set of control points facilitates user-controllable motion in 4D generation. 

\begin{figure}[t]
    \centering
    \includegraphics[]{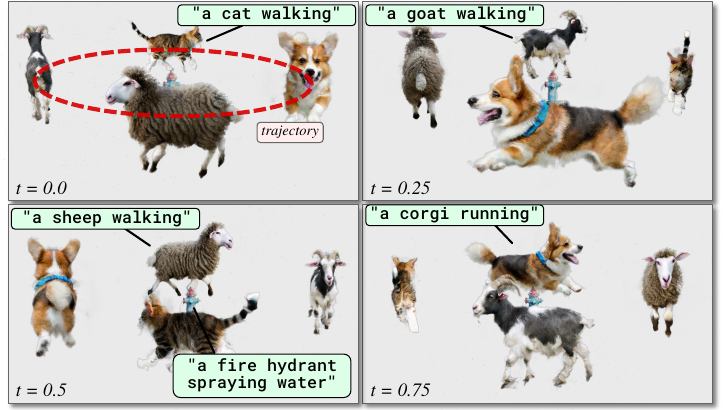}
    \caption{\textbf{Example of automated scene generation.} We use ChatGPT to automatically configure the layout of a scene from a text prompt. Specifically, ChatGPT takes the object descriptions as input and generates a configuration file that contains the trajectory for each object.}
    \label{fig:automated}
\end{figure}

\begin{figure}[t]
    \centering
    \includegraphics[width=0.8\textwidth]{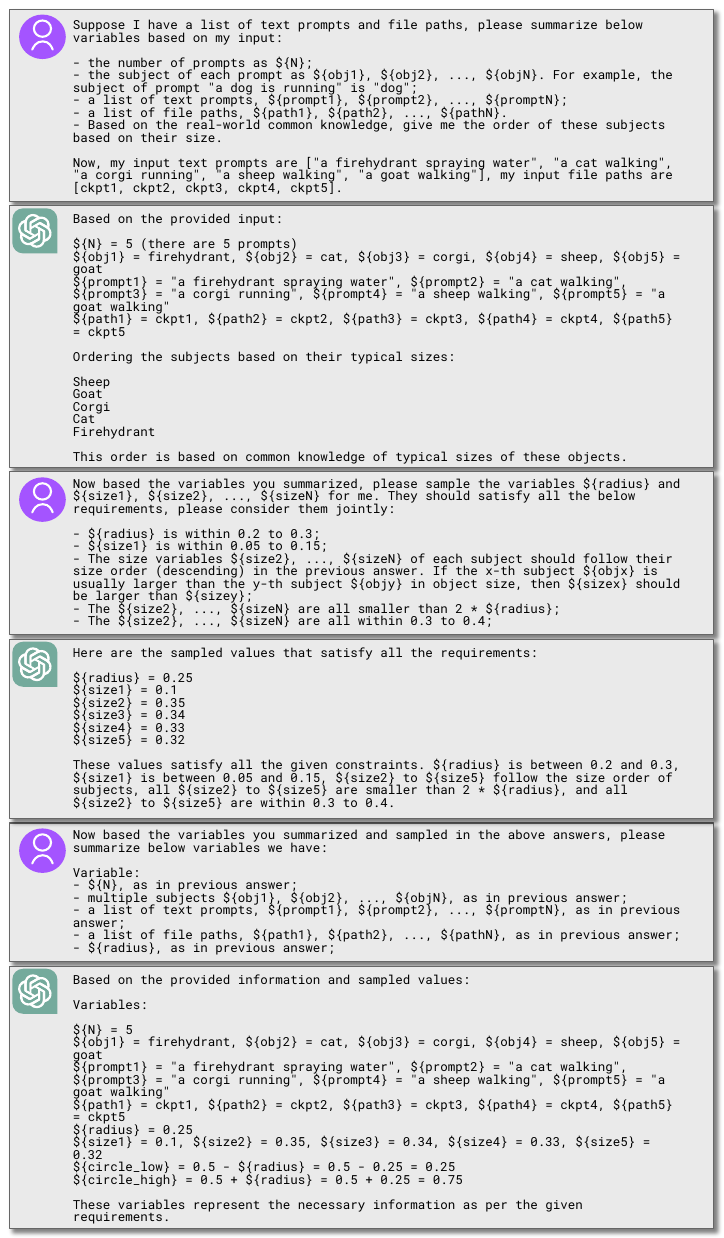}
    \caption{\textbf{Automatic scene generation using ChatGPT (part 1).} We provide a listing with the ChatGPT conversation used to generate the scene trajectories (continues in the next listing).}
    \label{fig:listing1}
\end{figure}

\begin{figure}[t]
    \centering
    \includegraphics[width=0.8\textwidth]{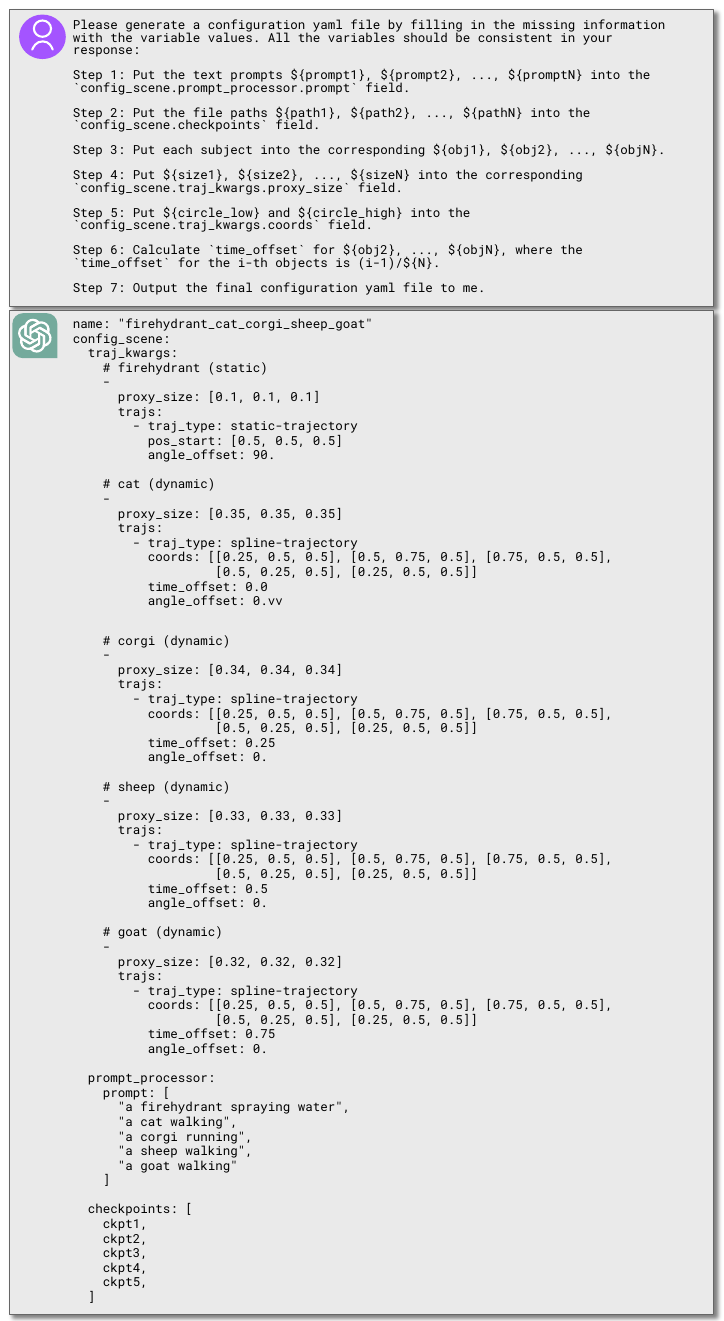}
    \caption{\textbf{Example of automated scene generation (part 2).} We provide a listing with the ChatGPT conversation used to generate the scene trajectories (continued from previous listing).}
    \label{fig:listing2}
\end{figure}

However, to demonstrate that our method can be automated, we explore the potential of large language models to fully automate the trajectory-conditioned text-to-4D generation process. 
Specifically, we use ChatGPT to specify the scene layout, including all object sizes and trajectory control points.  
The output of ChatGPT is a configuration file that can be input to our code to instantiate the 4D scene. 
We visualize the results of this procedure in Fig.~\ref{fig:automated}. 
While the result has some imperfections, e.g., the fire hydrant is too small compared to the other objects, we envision future work in this direction could enable high-fidelity scene layout and complex trajectory modeling for fully automated trajectory-conditioned text-to-4D generation. 
We show the conversation with ChatGPT used to generate the trajectories in Figs.~\ref{fig:listing1} and~\ref{fig:listing2}.

\begin{figure}[t]
  \centering
  \includegraphics[width=1.0\columnwidth]{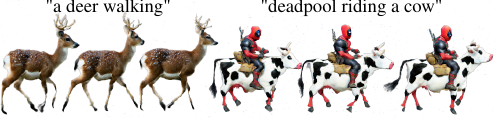} 
   \caption{Results using automated trajectory generation.}
   \label{fig:automated2}
\end{figure}

Moreover, we show single-object automated results in Fig.~\ref{fig:automated2}.
Improving automated trajectory generation and multi-object text-to-4D generation using, e.g., scene graphs~\cite{gao2023graphdreamer}, is an important direction for future work.

%% file: sec_supp/4_geometry.tex
\section{Geometry Visualization}
\label{sec:supp_geo_vis}

\begin{figure}
    \centering
    \includegraphics[]{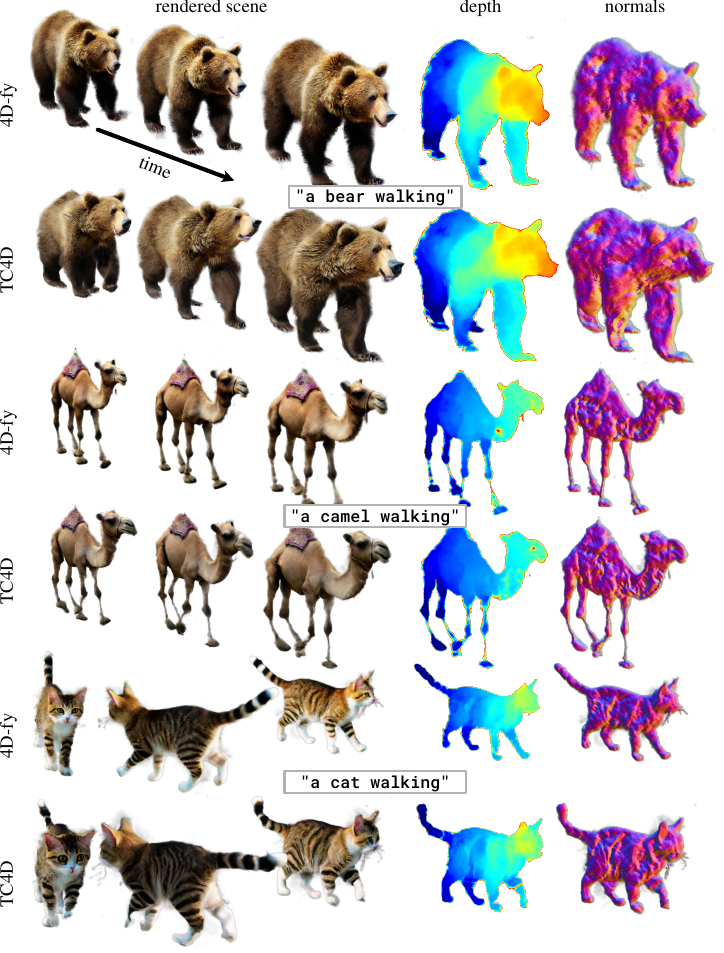}
    \caption{\textbf{Geometry visualization.} We show additional results comparing rendered images, depth maps, and normals from 4D-fy~\cite{bahmani20234d} and TC4D. The results from TC4D show improved motion with depth maps and normals that are of similar quality to 4D-fy.}
    \label{fig:geometry1}
\end{figure}

\begin{figure}
    \centering
    \includegraphics[]{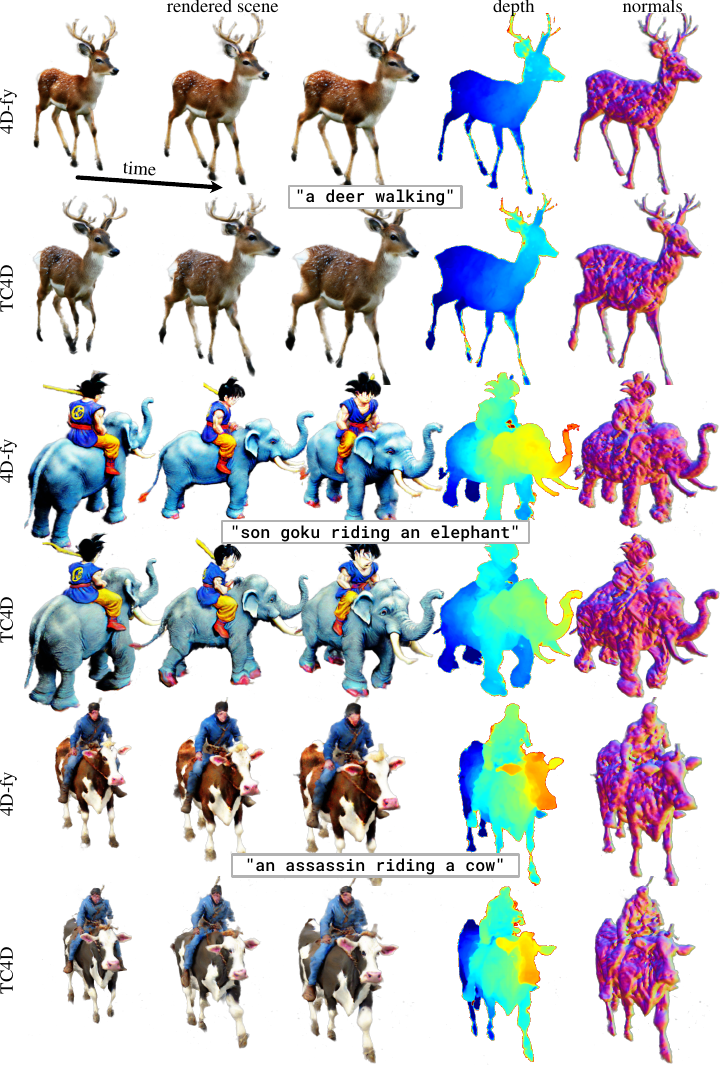}
    \vspace{-1em}
    \caption{\textbf{Geometry visualization.} We show additional results comparing rendered images, depth maps, and normals from 4D-fy~\cite{bahmani20234d} and TC4D. The results from TC4D show improved motion with depth maps and normals that are similar to 4D-fy.}
    \label{fig:geometry2}
\end{figure}

We show that TC4D generates scenes with comparable geometry to 4D-fy while introducing significantly more motion in Figs.~\ref{fig:geometry1} and ~\ref{fig:geometry2}. Note that the focus of this work is not to improve or provide high-quality geometry but introduce more realistic and larger-scale motion.

